\documentclass[letterpaper, 10pt, journal, twoside]{IEEEtran}
\usepackage{cite}
\usepackage{amsmath,amssymb,amsfonts}
\usepackage{graphicx}
\usepackage{textcomp}
\usepackage{xcolor}
\usepackage{url}
\usepackage{soul}
\usepackage{booktabs}
\usepackage{makecell}
\usepackage{multirow}
\usepackage{bm}
\usepackage[hidelinks]{hyperref}
\usepackage{eso-pic}

\begin{document}

\title{High-Speed, Scalable Sensor Readout for Dexterous Robotic Hands via Shift-Register Multiplexing}

\author{Jaehoon Kim$^{1}$,
Boshi An$^{1}$,
Lazaros Christoforidis$^{1}$,\\
Michail Papadakis$^{1}$,
Victor Kartsch$^{2}$,
and Robert K. Katzschmann$^{1,*}$%
\thanks{Manuscript received: May 11, 2026; Revised: August 5, 2026; Accepted: September 3, 2026.}%
\thanks{This paper was recommended for publication by Editor Lorenzo Jamone upon evaluation of the Associate Editor and Reviewers' comments. This work was supported by the SNSF Project Grant \#200021\_215489. Michail Papadakis received scholarship support from the Eugenides Foundation.}%
\thanks{$^{1}$Jaehoon Kim, Boshi An, Lazaros Christoforidis, Michail Papadakis, and Robert K. Katzschmann are with Soft Robotics Lab, IRIS, D-MAVT, ETH Zurich, Zurich, Switzerland.}%
\thanks{$^{2}$Victor Kartsch is with Integrated Systems Laboratory, ETH Zurich, Zurich, Switzerland.}%
\thanks{$^{*}$Corresponding author: \texttt{\href{mailto:rkk@ethz.ch}{rkk@ethz.ch}}}%
\thanks{Digital Object Identifier (DOI): \href{https://doi.org/10.1109/LRA.2026.3741779}{10.1109/LRA.2026.3741779}.}}

\markboth{IEEE Robotics and Automation Letters. Preprint Version. Accepted September, 2026}%
{Kim \MakeLowercase{\textit{et al.}}: High-Speed, Scalable Sensor Readout for Dexterous Robotic Hands}

\bstctlcite{IEEEtran:BSTcontrol}

\maketitle

\AddToShipoutPictureFG*{%
  \AtPageLowerLeft{%
    \put(\LenToUnit{\dimexpr1in+\hoffset+\oddsidemargin\relax},\LenToUnit{0.25in}){%
      \parbox[b]{\textwidth}{%
        \normalfont\fontsize{8}{8}\selectfont
        \copyright\ 2026 IEEE. Personal use of this material is permitted. Permission from IEEE must be obtained for all other uses, in any current or future media, including reprinting/republishing this material for advertising or promotional purposes, creating new collective works, for resale or redistribution to servers or lists, or reuse of any copyrighted component of this work in other works.\par
      }%
    }%
  }%
}

\begin{abstract}
Dexterous robotic hands require high-speed multimodal sensing across many degrees of freedom, yet existing readout architectures often impose trade-offs between sensor count, wiring complexity, and sampling bandwidth. This paper presents a scalable analog sensor readout architecture based on a serial-in parallel-out (SIPO) shift-register principle. The proposed architecture supports versatile integration of heterogeneous analog-output sensors, scalable expansion using only three signal lines between sensor modules, and fast, configurable sampling. We validate the approach on a tendon-driven robotic hand integrating 16 joint sensor modules and one four-channel tactile sensor module, enabling acquisition of 20 sensor channels at a full-scan rate of 1\,kHz, with stable operation up to 1.5\,kHz. Joint sensor characterization showed a maximum slope absolute percentage error (APE) of 0.446\% and sub-degree estimation error, indicating that the proposed readout system preserves sensing performance. 
For tactile sensing, LSTM-based models enabled real-time force estimation and contact-location classification at 1\,kHz.
System-level experiments showed that the joint sensors provided more accurate feedback than motor-based estimation during interaction. Closed-loop experiments further demonstrated improved joint trajectory tracking using joint-sensor feedback and force-controlled object pinching using tactile feedback.
The proposed architecture offers a practical path toward sensorized robotic hands for dexterous manipulation.
\end{abstract}

\begin{IEEEkeywords}
multifingered hands, force and tactile sensing, embedded systems for robotic and automation
\end{IEEEkeywords}

\begin{figure}[!t]
\centering
\includegraphics[width=\columnwidth]{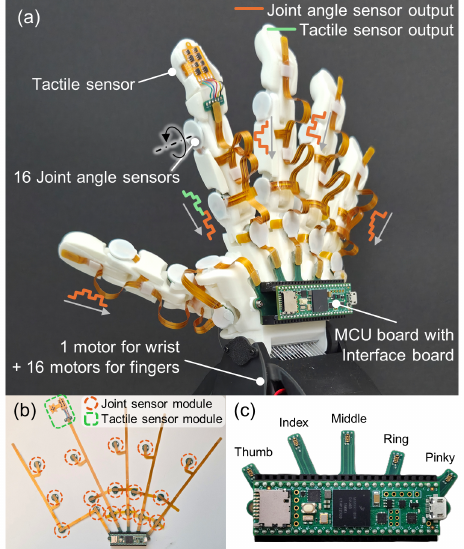}
\caption{Overview of the proposed sensorized robotic hand. (a) Integrated hand with 16 joint sensor modules and one tactile sensor module. The locations of the modules and the MCU board with the interface board are indicated. (b) Complete electrical hardware of the sensor system. (c) Custom MCU interface board with a Teensy 4.1 for pulse generation, clocking, and analog signal acquisition at a maximum full-scan rate of 1.5\,kHz.}
\label{fig:overall_system}
\end{figure}

\section{Introduction}

\IEEEPARstart{W}{ith} the growing prominence of humanoid robots and generalist robotic systems, research on robotic hands has become increasingly active. In such systems, rich sensory feedback is essential for interaction with the environment, making sensor integration and readout within the limited space of the hand a key design challenge.

Effective dexterous manipulation requires multimodal sensing \cite{xia_review_2022, saudabayev_sensors_2015}, including accurate proprioceptive sensing \cite{andrychowicz_learning_2020} and tactile information \cite{lee_dextouch_2024, hogan_tactile_2020}. In tendon-driven hands, motor encoder-based joint estimation can degrade due to tendon-transmission friction, compliance, and creep \cite{friedl_frcef_2015,liconti_leveraging_2024}, making dedicated joint sensors important. Tactile sensing is essential for human hand manipulation \cite{johansson_coding_2009}. In robotic systems, tactile sensing has also been shown to improve the performance of learning-based dexterous manipulation \cite{lee_dextouch_2024}.
Accordingly, robotic hands should integrate both proprioceptive and tactile sensors, and may additionally require other sensing modalities depending on the application \cite{chalon_spacehand_2015}. Therefore, sensor systems for robotic hands should flexibly support a wide variety of sensor types, which motivates the versatility addressed in this work.

Scalability is another key requirement for robotic hand sensor systems. Modern robotic hands have many degrees of freedom (DoF), such as 20 in the neoDavid hand \cite{wolf_neodavid_2025} and 17 in the ORCA hand \cite{christoph_orca_2025}, and the number of required sensor channels increases rapidly when multimodal sensing is considered. Because the interior of a robotic hand is already occupied by tendons and joint mechanisms, connecting many sensors with individual wires quickly becomes impractical. Furthermore, communication protocols such as I2C and SPI, which are widely used for digital sensor integration \cite{leens_introduction_2009}, impose practical constraints on scalability. I2C supports only a limited address space, and identical sensor devices often share fixed addresses, necessitating additional multiplexing circuitry for multi-sensor deployment. SPI, while avoiding address conflicts, typically requires a dedicated chip-select line for each sensor, rapidly exhausting the available GPIO resources of the microcontroller unit (MCU) as the sensor count increases.
Hierarchical digital architectures can support multisensory acquisition with reduced finger wiring. The DLR/HIT Hand II achieves this through local finger-level processing, but employs multiple DSPs and FPGAs across the fingers and palm in a multi-level processing and communication architecture \cite{liu_multisensory_2008}.

Fast and configurable sampling is another key requirement. Low-level feedback control in humanoid robotic systems is often implemented at rates on the order of 1\,kHz \cite{wolf_neodavid_2025, feng_optimization_2014}, motivating sensor readout at comparable rates.
In tactile sensing, human fingertip perception is highly sensitive to vibrotactile stimuli around 250~Hz, and tactile afferents can encode mechanical transients up to several hundred hertz \cite{johansson_coding_2009, verrillo_vibrotactile_1971}. To capture comparable dynamic tactile information in robotic tactile sensors, the Nyquist criterion requires sampling at least twice the relevant frequency. In practice, around 1\,kHz is desirable. Since sensing requirements vary across tasks, the readout architecture should support not only fast but also configurable sampling.

Motivated by these requirements and architectural trade-offs, this paper proposes a scalable sensor readout architecture based on a serial-in parallel-out (SIPO) shift register. The architecture centralizes ADC conversion and limits the local readout electronics of each sensing unit to a D flip-flop and an analog switch, thereby simplifying local integration. Sensor outputs are sequentially sampled over a shared analog line. The design is organized around three key properties: \textbf{versatility}, \textbf{scalability}, and \textbf{fast and configurable sampling}, as detailed in Section~\ref{sec:principle}.

\begin{figure}[!t]
\centering
\includegraphics[width=\columnwidth]{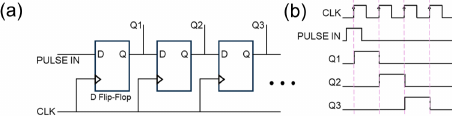}
\caption{Operating principle of the serial-in parallel-out (SIPO) shift register used for sensor readout. (a) Example circuit composed of three cascaded D flip-flops. (b) Timing diagram showing how an input pulse is sequentially shifted through Q1, Q2, and Q3 at each rising edge of the clock.}
\label{fig:principle}
\end{figure}

The main contributions of this paper are threefold. 
First, we propose a scalable analog sensor readout architecture that enables high-speed sampling and dense sensor integration with minimal wiring complexity.
Second, we present a sensorized robotic hand system integrating 16 joint sensor modules and one four-channel tactile sensor module for rich proprioceptive and tactile sensing (Fig.~\ref{fig:overall_system}). 
Third, we validate the proposed system through component-level characterization and system-level experiments. The component-level results show that the readout architecture has limited impact on sensing performance and that both the joint and tactile sensors provide accurate measurements. The system-level experiments further show that the joint sensors enable more accurate estimation during physical interaction than motor-based estimation, highlighting the need for dedicated joint sensing in tendon-driven hands. The closed-loop experiments further demonstrate the practical utility of the sensor measurements for joint trajectory control and force-regulated object pinching.

\section{System Design}

\subsection{Operating Principle of Sensor Readout}
\label{sec:principle}

\begin{figure}[!t]
\centering
\includegraphics[width=\columnwidth]{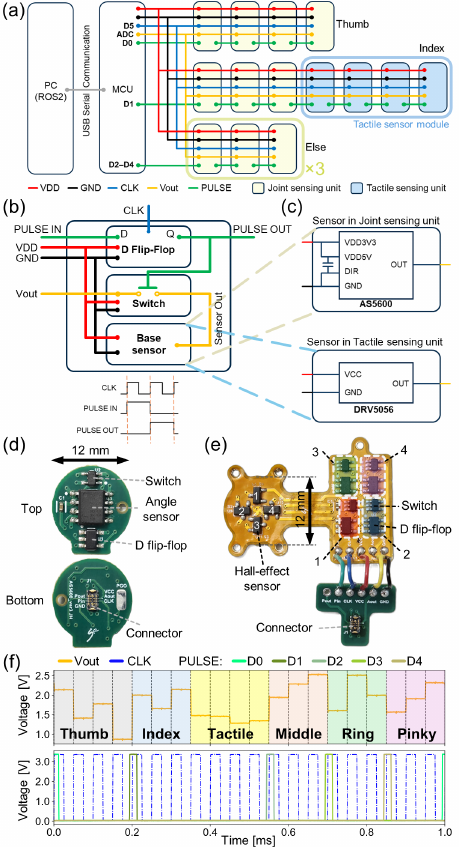}
\caption{Configuration and hardware implementation of the proposed sensor readout system. (a) Sensor system configuration of the robotic hand. (b) Circuit of a single sensing unit composed of a D flip-flop, an analog switch, and a base sensor; the bottom plot shows the unit-level timing diagram, in which the propagated pulse defines the time slot during which the switch connects the sensor output to the shared analog line. (c) Base sensors used for the joint (top) and tactile (bottom) sensing units. (d) Fabricated joint sensor module. (e) Fabricated tactile sensor module, with the four tactile sensing units labeled 1--4. (f) Oscilloscope measurements of the signal lines in the fully integrated system at a 1\,kHz full-scan rate, showing the multiplexed analog output (top) and the corresponding clock and pulse signals (bottom). The ADC read operation is initiated immediately after the falling edge of the clock and completed before the next rising edge.}
\label{fig:readout_config}
\end{figure}

In a robotic hand, each finger typically follows an open-chain structure, where links are sequentially connected, branching from the palm. 
We aimed to develop a circuit capable of multiplexing multiple sensors using a minimal number of components, while supporting minimal wiring, selective access to sensor groups, and fast sampling with simple control using a MCU.

The proposed circuit is based on a multiplexing architecture in which switches are controlled by digital pulses to sequentially connect sensors. Previous work \cite{nishino_tactile_2009} utilized delay circuits to propagate pulses and control switches. However, since such delay elements rely on analog components with parameter-dependent characteristics, they offer limited clock tunability and make it difficult to synchronize the sampling rate with the analog-to-digital converter (ADC).

To address these limitations, we designed a circuit based on the principle of a SIPO shift register. The circuit consists of D flip-flops and switches, enabling sequential readout of sensors connected in series. Fig.~\ref{fig:principle}(a) shows an example of a SIPO shift register with three outputs. When a single high pulse is applied to the input (D) of the first D flip-flop, the pulse propagates to subsequent flip-flops at each rising edge of the clock (Fig.~\ref{fig:principle}(b)). 
Each output (Q1--Q3) is connected to the control pin of a bilateral analog switch, so that the switches are activated sequentially. In each switch, one terminal is commonly connected to the ADC input, while the other terminal is connected to an individual sensor. This configuration allows each sensor to be sequentially connected to the ADC for one clock cycle.
The ADC performs sampling during the time slot in which each sensor is connected. This design offers several advantages:

\noindent\textbf{Versatility:}
Centralized ADC conversion enables the integration of heterogeneous analog-output sensors, including Hall-effect sensors and other analog-output sensor ICs.
\par
\noindent\textbf{Scalability:}
The architecture supports two forms of scaling: sequential extension and branching. Additional sensing units can be cascaded within each serial chain while the number of inter-module signal lines remains fixed at three, excluding VDD and GND. Multiple chains can also be arranged as parallel branches and selectively accessed through branch-specific pulse signals, enabling finger-wise sensor grouping.
\par
\noindent\textbf{Fast and configurable sampling:}
The full scan rate over all multiplexed sensors can be adjusted via the clock signal and synchronized with the MCU. Since sensors are sampled sequentially, the maximum scan rate is mainly limited by the ADC conversion time, enabling high-speed operation with fast ADCs.

\subsection{Sensor System Configuration}

\begin{figure*}[!t]
\centering
\includegraphics[width=\textwidth]{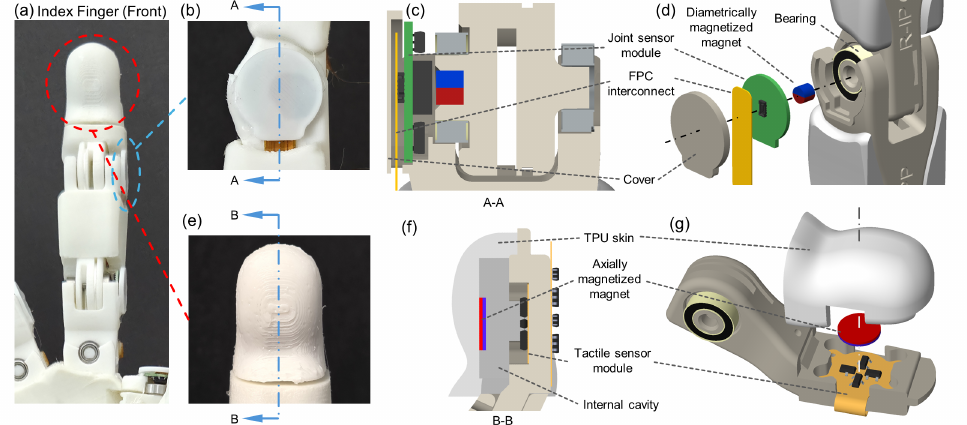}
\caption{Mechanical integration of the sensor system into the robotic hand. (a) Front view of the index finger showing the integrated sensing regions. (b) Enlarged view of the joint region. (c) Cross-sectional view A--A of the joint sensor structure. (d) Exploded view of the joint sensor assembly. (e) Enlarged view of the fingertip. (f) Cross-sectional view B--B of the tactile sensor structure. (g) Exploded view of the tactile sensor assembly.}
\label{fig:sensor_config}
\end{figure*}

To validate the three key advantages of the proposed sensor system, we integrated it into the ORCA hand, an open-source tendon-driven robotic hand with 16 finger DoF \cite{christoph_orca_2025}. The implemented system consists of 16 joint sensor modules and one four-channel tactile sensor module. 

Fig.~\ref{fig:readout_config}(a) shows the sensor system configuration. Clock and pulse signal generation and ADC-based sampling are performed by an MCU board (Teensy 4.1), which is mounted on a custom PCB designed for sensor interfacing and mechanical integration with the robotic hand (Fig.~\ref{fig:overall_system}(c)). The raw data is transmitted to a PC via USB serial and converted into joint angles, contact force, and contact-location on the PC with Robot Operating System 2 (ROS 2).

Joint angle and tactile sensors are designed in modular form. Each module consists of units that include a base sensor, a D flip-flop (SN74AUP1G79DBVR, Texas Instruments), and a bilateral analog switch (SN74LVC1G66DCKR, Texas Instruments) (Fig.~\ref{fig:readout_config}(b)).
The modules are connected to the MCU interface board via connectors on a flexible printed circuit (FPC) interconnect board. To minimize space, we use a PCB-to-PCB connector (BM29B0.6-4DS/2-0.35V, Hirose Electric Co., Ltd.) with a height below 1\,mm.

The joint sensor module consists of a single unit based on a magnetic rotary position sensor (AS5600, ams-OSRAM AG), which outputs an analog voltage proportional to the angle of a diametrically magnetized magnet placed directly above the sensor (Fig.~\ref{fig:readout_config}(c)). The module is implemented as a PCB with a protruding tab for anti-rotation (Fig.~\ref{fig:readout_config}(d)).

The tactile sensor module comprises four tactile sensing units implemented on a flexible printed circuit board (FPCB). The layout includes a circular sensing region and a rectangular circuit region separated by a component-free bending region, allowing the FPCB to be folded and fixed in the final structure (Fig.~\ref{fig:readout_config}(e)). Each unit uses a unipolar linear Hall-effect sensor (DRV5056A3QDBZR, Texas Instruments) as the base sensor (Fig.~\ref{fig:readout_config}(c)). The four sensors are evenly distributed around a common center and measure magnetic field variations induced by the motion of an axially magnetized magnet, enabling force estimation.

As shown in Fig.~\ref{fig:readout_config}(a), the thumb includes four joint sensor modules, the index finger includes three joint sensor modules and one tactile sensor module, and the remaining fingers each include three joint sensor modules connected in series. For each finger, the PULSE IN of the first module is connected to a dedicated digital I/O pin (D0--D4) of the MCU board, enabling selective finger-wise access.

Fig.~\ref{fig:readout_config}(f) shows oscilloscope measurements (Analog Discovery 3, Digilent) of the fully integrated system at a 1\,kHz full-scan rate over a 1\,ms window.
The upper plot shows the raw data from 20 sensor units sequentially connected to the Vout line, while the lower plot shows the corresponding clock and pulse signals. Each sensor time slot corresponds to one clock cycle, and the scan rate can be adjusted by changing the clock frequency. The maximum stable full scan rate for 20 sensor units was measured to be approximately 1.5\,kHz, corresponding to a clock frequency of 30\,kHz.
To quantify the power consumption of the local D flip-flop and analog switch, the four thumb joint modules were replaced with four sensor-free test boards, and the thumb-chain current was measured using a Power Profiler Kit II (Nordic Semiconductor) at a 1\,kHz full-scan rate, yielding an average power consumption of 0.571\,\textmu W per module.

\subsection{Mechanical Integration of the Sensor Modules}

As shown in Fig.~\ref{fig:overall_system}(a), the open-source hand was modified to accommodate the sensor system, with the MCU and FPC interconnect boards mounted on the dorsal side. The rigid structure was primarily 3D-printed in polylactic acid (PLA) and the soft skin in thermoplastic polyurethane (TPU). Fig.~\ref{fig:sensor_config} illustrates the integration of both module types.

The joint sensor module is mounted perpendicular to the joint axis, with the sensor at the axis center (Fig.~\ref{fig:sensor_config}(b)--(d)). A diametrically magnetized cylindrical magnet (NdFeB, $\varnothing$3\,mm $\times$ 2.5\,mm) is embedded at the center of the shaft on the inner side of the bearing, directly above the sensor, enabling angle measurement. The tactile sensor module is folded so that its circular region attaches to the front of the fingertip and its rectangular region to the back, with a TPU skin on top (Fig.~\ref{fig:sensor_config}(e)--(g)). An axially magnetized disc magnet (NdFeB, $\varnothing$8\,mm $\times$ 1\,mm) on the inner surface of the skin displaces within an internal cavity under external force, and the resulting magnetic-field variation is measured by the four Hall-effect sensors.

\subsection{Deep Learning Models for Force Estimation and Contact-Location Classification Using a Tactile Sensor}
\label{sec:lstm}

The proposed sensor system provides synchronized four-channel tactile measurements at 1\,kHz, enabling temporal tactile estimation in real time. As a system-level demonstration of the versatility of the proposed readout, we integrate the four-channel tactile module and use it for contact-force estimation and contact-location classification. Following prior work on temporal tactile sensing \cite{han_use_2018}, we adopt LSTM-based models.
The two tasks are handled by separate models that share a backbone of two stacked LSTM layers with 32 and 12 hidden units, each followed by dropout of 0.2. The force-estimation model takes a sequence of length 20 from the four Hall-effect sensor channels and outputs a scalar force through a fully connected layer with ReLU activation. The contact-location model takes a sequence of length 80 from the same channels and outputs five contact states (0: no contact, 1--4: contact locations) through a fully connected layer with softmax activation. The models are trained independently using mean squared error for force estimation and categorical cross-entropy for classification.

\section{Results and Discussion}


\begin{figure}[!t]
\centering
\includegraphics[width=\columnwidth]{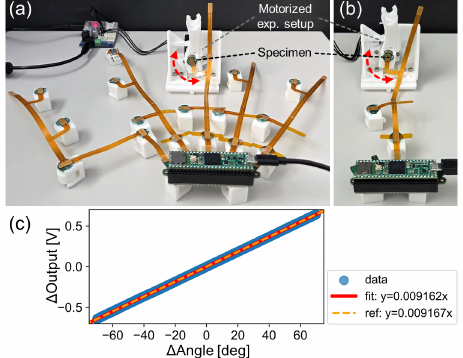}
\caption{Experimental setup for characterization of the joint sensor module and readout system. Motorized experimental setup with (a) the full sensor system configuration, consisting of 16 joint sensor modules and one tactile sensor module (four Hall-effect sensors), and (b) the simplified configuration, consisting of only the three joint sensor modules of the middle finger. (c) Representative sensor output as a function of joint angle change from one of the five trials under the experimental condition corresponding to index 2 in Table~\ref{tab:noise-table}, together with the fitted slope and the reference slope.}
\label{fig:char_joint}
\end{figure}

\subsection{Characterization of the Joint Sensor Module and Readout System}

\textbf{Experimental Setup and Evaluation Metrics}:
To evaluate the joint sensor modules, a motorized setup was developed to record the ground-truth angle and sensor output (Fig.~\ref{fig:char_joint}(a), (b)). The setup emulates a single joint of the robotic hand, driven directly by a motor (XC330-T288-T, ROBOTIS Co., Ltd.).

The joint was actuated sinusoidally from $-70^\circ$ to $70^\circ$ with a 20\,s period for five cycles.
During operation, the ground-truth angle was obtained from the motor's internal absolute encoder (12-bit resolution over $360^\circ$), while the output of the sensor module mounted on the joint was recorded.
The tested specimen was the third module in the middle-finger serial chain.
For the remaining joint sensor modules, diametrically magnetized magnets were placed at arbitrary orientations to generate constant voltage outputs.

To assess the influence of the sensor readout system on sensing performance, experiments were conducted under various configurations.
Scan rate effects were evaluated at 0.5, 1.0, and 1.5\,kHz using specimen A in the full configuration of Fig.~\ref{fig:char_joint}(a), comprising 16 joint sensor modules and one tactile sensor module.
Sensor-coupling effects were evaluated at 1.0\,kHz using the same specimen in the simplified configuration of Fig.~\ref{fig:char_joint}(b), containing only the three middle-finger joint sensor modules.
Finally, to evaluate specimen-to-specimen variation, the 1.0\,kHz experiment in the full configuration was repeated for specimens B--D.

Because zeroing is performed in the robotic hand system and magnet alignment cannot be perfectly matched across joints, the analysis is based on incremental values, i.e., $\Delta V$ and $\Delta \theta$. For all metric computations, the sensor output was temporally aligned to the ground-truth angle using interpolation.
Each experimental configuration was repeated five times, and the reported metrics correspond to the mean and sample standard deviation across the five trials.
The metrics are defined below.

Since independent calibration of each joint sensor module is impractical after integration in the space-constrained hand, angle estimation uses the linear angle-to-voltage relationship specified in the AS5600 datasheet. The corresponding slope represents the sensor sensitivity:
\begin{equation}
a_{\mathrm{ref}} = \frac{V_{\mathrm{DD}}}{360} = \frac{3.3}{360} = 0.009167\ \mathrm{V/deg}.
\end{equation}

The absolute percentage error (APE) of the slope is computed by comparing this reference slope with the slope obtained from a least-squares fit constrained to pass through the origin:
\begin{equation}
\mathrm{APE}_{\mathrm{slope}} (\%) =
\left|
\frac{a_{\mathrm{fit}} - a_{\mathrm{ref}}}{a_{\mathrm{ref}}}
\right| \times 100.
\end{equation}

The estimation error at sample \(i\) is computed as
\begin{equation}
e_i =
\frac{\Delta V_i}{a_{\mathrm{ref}}}
-
\Delta \theta_{\mathrm{true},i}.
\end{equation}

The root mean square error (RMSE$_{\mathrm{ref}}$) and the standard deviation (STD$_{\mathrm{error}}$) of the error are computed from $e_i$. 
The results are summarized in Table~\ref{tab:noise-table}.

\textbf{Results}:
Table~\ref{tab:noise-table} shows a maximum slope APE of 0.446\% for specimen B in the full sensor configuration.
Since specimens B and C showed larger APEs than specimens A and D under the same readout configuration, this variation is more likely attributable to specimen-to-specimen differences, including variation in the base sensors and assembly tolerances, rather than to the readout architecture itself.

Assuming a linear sensor model $\Delta V = a \Delta \theta$ and estimating the joint angle change using the reference slope as $\Delta \theta_{\mathrm{est}} = \Delta V / a_{\mathrm{ref}}$, the resulting relative estimation error becomes
\begin{equation}
\frac{\Delta \theta_{\mathrm{est}} - \Delta \theta}{\Delta \theta} = \frac{a - a_{\mathrm{ref}}}{a_{\mathrm{ref}}}.
\end{equation}
Thus, the slope APE directly corresponds to the relative scale-factor error in the estimated joint angle change. 
In our case, the maximum APE of 0.446\% corresponds to a worst-case scale-factor-induced absolute error of approximately $0.31^\circ$ at the maximum tested angle magnitude of $70^\circ$.

To evaluate the overall sensor performance beyond slope error alone, we further analyze the RMSE$_{\mathrm{ref}}$ and STD$_{\mathrm{error}}$ summarized in Table~\ref{tab:noise-table}. The RMSE$_{\mathrm{ref}}$ ranges from $0.233^\circ$ to $0.368^\circ$, and the STD$_{\mathrm{error}}$ ranges from $0.227^\circ$ to $0.343^\circ$ across different experimental conditions. The similar magnitudes of these two metrics suggest that any mean bias component is small compared with the overall error variability. In all cases, both metrics remain below $0.37^\circ$, demonstrating stable sub-degree angle estimation using the reference slope.

Overall, the results show that the proposed sensor readout system has limited impact on sensing performance, while the joint sensor characterization demonstrates sub-degree estimation error across all tested conditions. In addition, stable operation was maintained across all tested scan rates up to 1.5\,kHz, demonstrating the robustness of the proposed readout system under high-speed acquisition conditions.

\begin{table}[t]
\centering
\fontsize{7.5}{8}\selectfont
\caption{Characterization Results of the Joint Sensor Module and Readout System.}
\label{tab:noise-table}
\setlength{\tabcolsep}{3pt}
\renewcommand{\arraystretch}{1.15}
\begin{tabular}{c c c c | c c c}
\toprule
\makecell{Idx.}
& \makecell{Spec.} 
& \makecell{SR \\ {[kHz]}} 
& \makecell{Sensor \\ Comb.} 
& \makecell{APE$_{\mathrm{slope}}$ \\ {[\%]}} 
& \makecell{RMSE$_{\mathrm{ref}}$ \\ {[deg]}} 
& \makecell{STD$_{\mathrm{error}}$ \\ {[deg]}} \\
\midrule

1 & \multirow{4}{*}{A}
& 0.5  & \multirow{3}{*}{\makecell[c]{16 J \\ + \\ 4 H}} 
& 0.120 (0.0101) & 0.261 (0.0529) & 0.232 (0.0038) \\
2 & & 1.0 &  
& 0.065 (0.0095) & 0.234 (0.0005) & 0.227 (0.0014) \\
3 & & 1.5 &  
& 0.063 (0.0205) & 0.237 (0.0030) & 0.231 (0.0028) \\
\cmidrule(lr){3-4}
4 & & \multirow{4}{*}{1.0} & 3 J 
& 0.066 (0.0104) & 0.233 (0.0020) & 0.228 (0.0023) \\

\cmidrule(lr){2-2}\cmidrule(lr){4-4}
5 & B &  & \multirow{3}{*}{\makecell[c]{16 J \\ + \\ 4 H}} 
& 0.446 (0.0076) & 0.368 (0.0573) & 0.343 (0.0041) \\
6 & C &  &  
& 0.434 (0.0131) & 0.356 (0.0886) & 0.316 (0.0077) \\
7 & D &  &  
& 0.145 (0.0167) & 0.259 (0.0025) & 0.258 (0.0020) \\

\bottomrule
\end{tabular}

\vspace{1mm}
\parbox{\columnwidth}{\raggedright \footnotesize Each condition was repeated five times. Reported values are the mean of each metric, with sample standard deviations given in parentheses.

Spec.: specimen; SR: scan rate; Comb.: combination; J: joint sensors; H: Hall-effect sensors.}
\end{table}


\subsection{Characterization of the Tactile Sensor Module}

\begin{figure}[!t]
\centering
\includegraphics[width=\columnwidth]{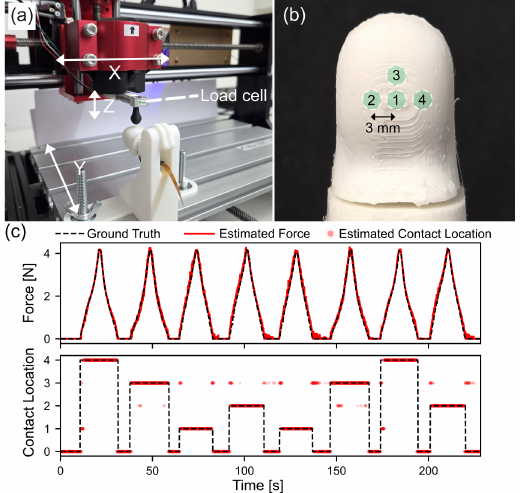}
\caption{Characterization of the tactile sensor module. (a) CNC-based motorized experimental setup for data collection using a load cell and a 3D-printed indenter. (b) Predefined contact-locations used for dataset collection. 
(c) Test set result, showing the estimated force and contact-location together with the corresponding ground truth.}
\label{fig:char_tactile}
\end{figure}

The tactile sensor module is used to estimate both force and contact-location over five classes (0: no contact, 1--4: contact locations) using LSTM-based models. 

\textbf{Experimental Setup}:
Experiments were conducted using a 3-axis CNC machine (TTC3018 Pro, Twotrees) equipped with a load cell (capacity: 500 g, TAL221, HT Sensor Technology) and a 3D-printed indenter (Fig.~\ref{fig:char_tactile}(a)).
The CNC was programmed to move the indenter to predefined $(x, y)$ contact-locations. At each visit to a location, a loading-unloading cycle up to the predefined force was performed along the z-axis at a constant feed rate of 0.3\,mm/s.
During the experiments, ground-truth force from the load cell and sensor system data were recorded. The ground-truth contact labels were assigned offline based on the experimental protocol and force measurements.

\textbf{Data Collection}:
Separate training, validation, and test datasets were collected over the four labeled contact-locations in Fig.~\ref{fig:char_tactile}(b), with loading and unloading up to 4\,N at each location. The training and validation sets used six and two sequences, respectively, pressing locations 1--4 in a fixed order. The test set used two sequences with randomized contact order to evaluate generalization to unseen temporal sequences.

\textbf{Results}:
For the test dataset, the temporally aligned model predictions were compared with the ground-truth force and contact labels (Fig.~\ref{fig:char_tactile}(c)).

For force estimation, the RMSE over the entire test set was 0.125\,N. To exclude the effect of near-zero-force regions, an additional evaluation was performed for samples with force greater than 0.05\,N, resulting in an RMSE of 0.148\,N. The slightly higher error reflects the increased difficulty of force estimation during contact. Considering the full-scale force range of 4\,N, the observed error remains below 4\% of full scale, demonstrating accurate force estimation.
For contact-location classification, the classification accuracy on the test set was 93.4\%. 
Together, these results indicate the suitability of the tactile signals acquired through the proposed readout architecture for downstream inference tasks, as illustrated by force estimation and contact-location classification.

For deployment, the trained models were integrated into a ROS 2 package, and real-time inference at 1\,kHz was verified on a PC running Ubuntu 24.04.3 LTS with an Intel Core i9-13900H CPU, 32\,GB RAM, and an NVIDIA GeForce RTX 4070 Laptop GPU (8\,GB).


\subsection{System-Level Evaluation of the Sensorized Robotic Hand}

\begin{figure}[!t]
\centering
\includegraphics[width=\columnwidth]{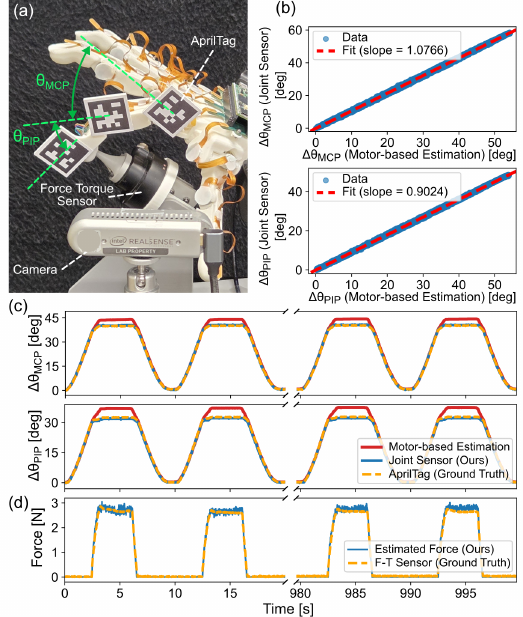}
\caption{Evaluation of the proposed sensorized robotic hand. (a) Experimental setup with vision-based joint angle measurement and force torque (F-T) sensing during fingertip interaction. (b) Calibration result for compensating the motor-based joint estimation, showing the relationship between the motor-based estimation and the joint sensor output for the MCP and PIP joints. (c) Joint angle estimation during interaction, showing the motor-based estimation, joint sensor output, and vision-based ground truth for the MCP and PIP joints. (d) Tactile force estimation result compared with the F-T sensor measurement. In (c) and (d), only the first two and last two cycles of the 100-cycle experiment are shown.}
\label{fig:sensor_orca}
\end{figure}

To validate the system-level performance of the proposed system, experiments were conducted using a tendon-driven robotic hand equipped with joint sensor modules and a fingertip tactile sensor module. This evaluation aims to demonstrate the necessity of joint sensing in tendon-driven mechanisms and the accuracy of tactile force estimation during interaction.

\textbf{Experimental Setup}:
The experimental setup is shown in Fig.~\ref{fig:sensor_orca}(a), where the index finger of the robotic hand was evaluated. AprilTag markers attached to each link were tracked by a camera (D435, Intel RealSense) to obtain ground-truth joint angles, which were recorded together with the joint angles estimated from the proposed joint sensor modules and from the motor-based estimation (based on the ORCA hand driver). A six-axis force torque sensor (F-T sensor, MiniONE Pro, Bota Systems AG) was positioned to obstruct the fingertip trajectory and induce contact interactions. Its three-axis force magnitude was used as the ground truth for force estimation, while the tactile sensor module provided the tactile data. Camera data, F-T measurements, sensor system data, and motor-based estimation were recorded at 60\,Hz, 100\,Hz, 1\,kHz, and 99\,Hz, respectively, using ROS 2 rosbag.

Because the ORCA hand is tendon-driven, tendon re-tensioning and calibration for motor-based estimation based on the joint range of motion were performed beforehand. Nevertheless, 3D-printed structural inaccuracies and transmission compliance can leave systematic errors in this estimation. To ensure a fair comparison, the motor-based estimation was additionally calibrated against the joint sensor output. In a free-space condition, the MCP and PIP joints were driven by a sinusoidal command between $0^\circ$ and $56.7^\circ$ (10\,s period, 10 cycles). Because of residual errors in the prior calibration, the actual joint motion deviated from this commanded range. A linear fit between the motor-based estimation and the joint sensor output (Fig.~\ref{fig:sensor_orca}(b)) then yielded a compensation slope, which was applied to the motor-based estimation in the subsequent experiment. Finally, the same command was applied for 100 cycles while interacting with the F-T sensor. All joint angles were compared in incremental ($\Delta$) values relative to their initial states. Because the modalities had different sampling rates, the estimates were temporally aligned to the corresponding ground truth.

\textbf{Results}:
The joint angle estimation results are shown in Fig.~\ref{fig:sensor_orca}(c). During interaction, substantial errors appear in the motor-based estimation, whereas the joint sensor measurements remain consistent with the ground truth.
RMSEs were computed separately for each of the 100 cycles. For the MCP joint, the resulting per-cycle RMSEs had means (standard deviations) of $0.387^\circ$ ($0.011^\circ$) for the joint sensor and $2.343^\circ$ ($0.014^\circ$) for the motor-based estimation. For the PIP joint, the corresponding values were $0.598^\circ$ ($0.014^\circ$) and $2.599^\circ$ ($0.044^\circ$), respectively.
These results show that compliance and transmission uncertainties in tendon-driven mechanisms can substantially degrade motor-based estimation during interaction, making dedicated joint sensing essential for accurate feedback.

The tactile force estimation result is shown in Fig.~\ref{fig:sensor_orca}(d). The estimated force closely follows the ground-truth force.
Across the 100 cycles, the per-cycle force-estimation RMSEs had means (standard deviations) of $0.090$\,N ($0.014$\,N) for all samples and $0.136$\,N ($0.023$\,N) for samples with force greater than $0.05$\,N.

A limitation of the current setup is that the F-T sensor was tilted and equipped with an indenter to emulate the dataset collection conditions, resulting in predominantly normal force components. As a result, shear-force estimation was not evaluated in the present experiment. 

\subsection{Sensor-Based Control and Grasping Demonstrations}

\begin{figure}[t]
    \centering
    \includegraphics[width=\columnwidth]{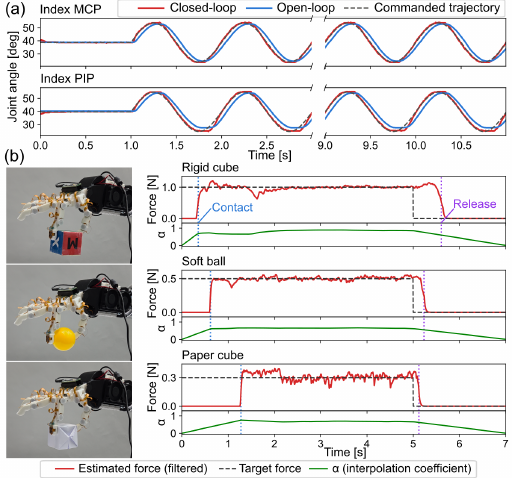}
    \caption{Closed-loop control experiments. 
    (a) Joint trajectory tracking under closed-loop and open-loop control. All 16 joints were commanded to follow 10 cycles of 1\,Hz sinusoidal motion after a 1\,s pause. The trajectories of two representative index-finger joints are shown for the initial pause, the first two cycles, and the final two cycles using a broken time axis. 
    (b) Tactile-sensor-based pinching of a rigid cube, a soft ball, and a paper cube, showing the force and interpolation-coefficient profiles for different target grasping forces.
    The contact and release points denote the onset and end, respectively, of the period during which the filtered force estimate is at or above 50\% of the maximum target force.
    }
    \label{fig:sine_tracking}
\end{figure}

\textbf{Closed-Loop Joint Trajectory Tracking:} To illustrate how the proposed sensor system can benefit downstream tasks, we conducted closed-loop control experiments using joint-sensor feedback. After a 1\,s pause, all 16 joints were commanded to follow 10 cycles of 1\,Hz sinusoidal trajectories spanning their respective motion ranges while maintaining a safety margin. Fig.~\ref{fig:sine_tracking}(a) shows the trajectories of two representative index-finger joints.
The controllers were run at 200\,Hz.
The closed-loop control law is defined as
\begin{equation}
\label{eq:controller}
\bm{p}^{\mathrm{motor}}_{T}
=
\operatorname{Motor}\left(
\bm{p}^{\mathrm{cmd}}_{T}
+
k_I^{\mathrm{pos}}
\int_{0}^{T}
\left(
\bm{p}^{\mathrm{cmd}}_{t}
-
\bm{p}^{\mathrm{sensor}}_{t}
\right)
\,\mathrm{d}t
\right).
\end{equation}
Here, $\bm{p}^{\mathrm{cmd}}_t$, $\bm{p}^{\mathrm{sensor}}_t$, and $\bm{p}^{\mathrm{motor}}_t$ denote the commanded joint-position, sensor-measured joint-position, and motor-position command vectors, respectively. 
The mapping $\operatorname{Motor}:\mathbb{R}^{16}\to\mathbb{R}^{16}$ converts joint-position commands to motor-position commands based on the hand's transmission kinematics, gear ratios, and motor direction conventions.
The open-loop controller is obtained by setting $k_I^{\mathrm{pos}}=0$.
For each joint, the tracking RMSE between the commanded joint positions and the joint positions measured by the joint sensors was computed over the full trajectory. Across the 16 joints, the closed-loop and open-loop controllers yielded RMSEs of $0.92^\circ~(0.24^\circ)$ and $3.13^\circ~(0.39^\circ)$, respectively, reported as mean (standard deviation).

\textbf{Tactile-Based Pinching:} 
The index finger and thumb were used to pinch a rigid cube, a soft ball, and a paper cube, which differed in shape and compliance (Fig.~\ref{fig:sine_tracking}(b)).
The hand was mounted on a Franka Panda robotic arm to reach and lift each object.
We defined two reference joint-space configurations, $\bm{p}^{\mathrm{open}}, \bm{p}^{\mathrm{full}}\in\mathbb{R}^{16}$, representing the open and fully pinched poses, respectively.
The commanded pinching pose at time $T$ was interpolated between the two reference configurations as $\bm{p}^{\mathrm{cmd}}_T=\alpha_T\bm{p}^{\mathrm{full}}+(1-\alpha_T)\bm{p}^{\mathrm{open}}$.
A clipped integral controller determines the interpolation coefficient $\alpha_T$ from the force error:
\begin{equation}
\alpha_T
=
\operatorname{clip}
\left(
k_I^{\mathrm{force}}
\int_0^T
\left(
f_t^{\mathrm{des}}
-
f_t^{\mathrm{est}}
\right)
\,\mathrm{d}t,
0,1
\right),
\end{equation}
where $f^{\mathrm{des}}$ is the desired pinching force, $f^{\mathrm{est}}$ is the LSTM-based force estimate in Sec.~\ref{sec:lstm} after exponential moving-average filtering (coefficient $0.8$), and $\operatorname{clip}(\cdot,0,1)$ constrains $\alpha_T$ to $[0,1]$. The interpolated command $\bm{p}^{\mathrm{cmd}}_T$ was computed at 200\,Hz and passed to the closed-loop controller in Eq.~\eqref{eq:controller}.

During the steady-state period from 3.0 to 4.8\,s, the RMSEs between the unfiltered force estimates and the corresponding target forces were $0.038$\,N, $0.032$\,N, and $0.055$\,N for the rigid cube, soft ball, and paper cube, respectively.
The hand successfully pinched all three objects.

\section{Conclusion}
This work presents a SIPO shift-register-based analog readout architecture that centralizes ADC conversion and maintains three inter-module signal lines within each serial sensor chain. By limiting the local readout electronics of each sensing unit to a D flip-flop and an analog switch, the architecture enables compact integration of heterogeneous analog-output sensors and kilohertz-rate acquisition in a sensorized robotic hand.
The implemented system acquired 20 channels from 16 joint sensor modules and one four-channel tactile module at a full-scan rate of 1\,kHz, with stable operation up to 1.5\,kHz. Characterization demonstrated sub-degree joint sensing, tactile force and contact-location estimation, and more accurate joint-angle estimation than motor-based estimation during physical interaction. The joint and tactile measurements were further used for closed-loop trajectory tracking and object pinching.

The current validation is limited to 20 sensing units and tactile sensing on a single fingertip under primarily normal-contact conditions. Moreover, sequential acquisition reduces the achievable full-scan rate as the channel count increases.

Future work will optimize the hardware and firmware for larger sensor counts and characterize circuit-level signal integrity and timing behavior, enabling systematic comparison with alternative sensor readout architectures based on serial communication protocols. Tactile sensing capabilities will be extended to resolve normal and shear interactions across varied contact geometries, and the benefits of kilohertz-rate tactile acquisition will be evaluated in dynamic contact and tactile-based manipulation tasks.
For joint sensing, the readout system will be integrated directly with low-level motor drivers to enable kilohertz-rate feedback control, which will be evaluated in dynamic manipulation tasks.
The supporting hardware and software will be released as open-source resources as the system matures.

\bibliographystyle{IEEEtran}
\bibliography{IEEEabrv,refs,IEEEtran_etal_control}

\end{document}